\documentclass[conference]{IEEEtran}
\IEEEoverridecommandlockouts    

\usepackage{lipsum}
\usepackage[ruled,vlined]{algorithm2e}

\usepackage{cite}
\usepackage{amsmath,amssymb,amsfonts}
\usepackage{algorithmic}
\usepackage{graphicx}
\usepackage{textcomp}
\usepackage{xcolor}
\usepackage{booktabs}
\usepackage{microtype}
\usepackage{multirow}
\usepackage{subfig}
\usepackage{caption}
\usepackage{wrapfig}
\usepackage[hidelinks]{hyperref}

\title{\LARGE \bf ChatBEV: Empowering Traffic Scene Understanding and Simulation via Vision-Language Model
}

\author{
	\parbox{\textwidth}{%
		\centering
		Qingyao Xu$^{1}$, Ya Zhang$^{1,2}$, Yanfeng Wang$^{1,2}$, Siheng Chen$^{1,2*}$%
	}%
	\thanks{$^{1}$Shanghai Jiao Tong University, Shanghai, China 
		{\tt\small {\{xuqingyao, ya\_zhang, wangyanfeng622, sihengc\}}@sjtu.edu.cn}%
	}%
    \thanks{$^{2}$Shanghai AI Laboratory, Shanghai, China}
    \thanks{$*$ Corresponding author}
}

\begin{document}
	
	\maketitle
	\thispagestyle{empty}
	\pagestyle{empty}
	
	\begin{abstract}
		Comprehensive traffic scene understanding is a foundational capability for Intelligent Transportation Systems (ITS) underpinning applications such as traffic simulation. While VisionLanguage Models (VLMs) have demonstrated strong reasoning potential, their application to Bird’s-Eye View (BEV) maps in traffic contexts remains limited by narrow task definitions and scarce annotated data. We introduce ChatBEV-QA, a large-scale BEV VQA benchmark of 137K+ QA pairs, designed to evaluate global scene understanding, vehicle-lane interactions, and vehiclevehicle interactions within complex traffic environments. Building on this, we fine-tune ChatBEV, a specialized VLM that accurately interprets diverse scene understanding queries from BEV maps. To demonstrate downstream utility in ITS applications, we integrate ChatBEV into a language-guided traffic simulation framework. Its global understanding and navigation reasoning provide crucial context-aware guidance, reducing trajectory displacement error by up to 20.9\% and scenario collision rates by up to 37.9\% over text-only baselines. \emph{ Code is available at \href{https://github.com/xuqingyao/ChatBEV}{https://github.com/xuqingyao/ChatBEV}.} 
\end{abstract}

\section{Introduction}
\label{sec:intro}

Traffic scene understanding~\cite{pan2018spatial,geiger20133d} is a fundamental task aims perceiving and interpreting the surrounding traffic environment. 
It serves as the cornerstone for intelligent transportation systems and autonomous driving, enabling informed decision-making and safe vehicle operation. In intelligent transportation systems, it optimizes traffic flow and prevents accidents by analyzing lane structures, traffic conditions, and vehicle interactions. For autonomous driving, it facilitates safer navigation by enhancing real-time motion planning. It also provides context-aware guidance for scene simulation, leading to more precise and controllable generation results.

Bird’s-Eye View (BEV) maps, with
their accurate representation of road geometry, lane boundaries, and vehicle positions, serve as a natural bridge between frontend perception and back-end ITS tasks. Given BEV’s central role, enabling semantic scene understanding of these BEV maps is a critical and timely challenge. This work thus focuses on Bird’s-Eye View (BEV)-based traffic scene understanding and its downstream application in controllable traffic simulation.

Traditional traffic scene understanding methods~\cite{li2020end, roddick2020predicting, can2022understanding} often focus on isolated objectives, lacking the holistic reasoning needed for comprehensive interpretation. Vision-Language Models (VLMs)~\cite{liu2024visual,achiam2023gpt} offer exceptional visual reasoning and the ability to answer free-form queries, making them a natural fit for BEV-based traffic understanding. However, their application to BEV traffic understanding remains underdeveloped, as existing benchmarks provide neither the task coverage nor the scale required to adapt VLMs to the structured geometry of BEV maps. Specifically, benchmarks like Talk2BEV~\cite{choudhary2023talk2bev} and MAPLM~\cite{cao2024maplm} are limited in scale or focus solely on object attributes, omitting the critical vehicle-lane and vehicle-vehicle interactions essential for traffic safety and conflict detection.

\begin{figure}[t]
  \centering
  \includegraphics[width=0.9\linewidth]{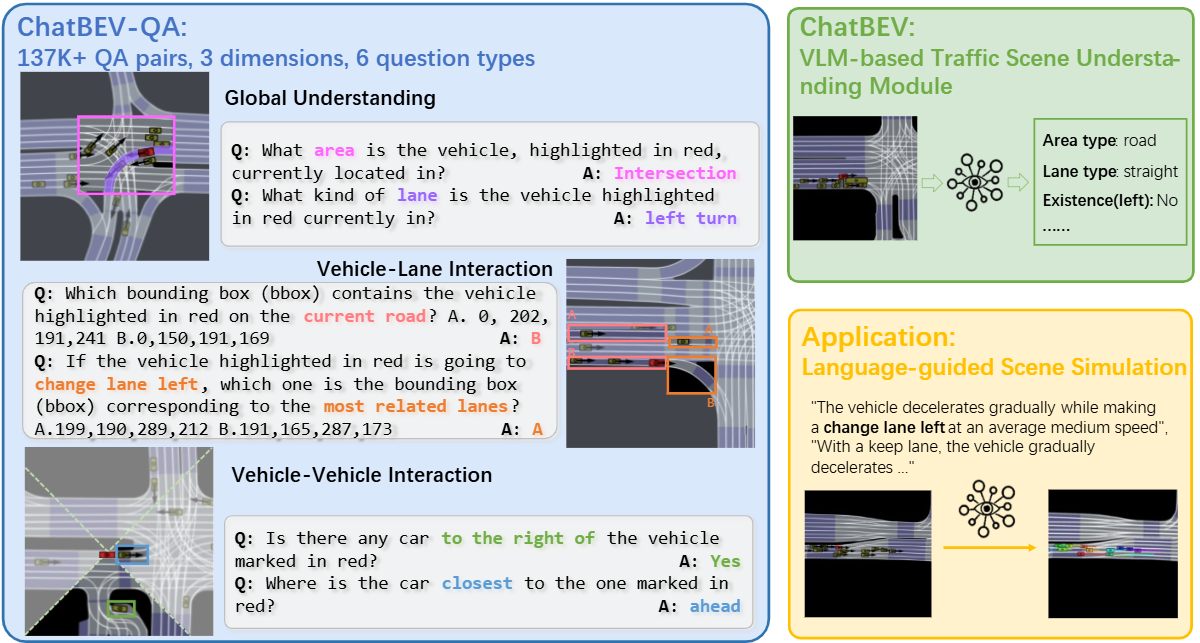}
  \caption{ChatBEV-QA is a scalable BEV VQA benchmark for comprehensive traffic scene understanding. Fine-tuned on ChatBEV-QA, ChatBEV provides high-level scene interpretation that directly supports downstream applications such as language-guided traffic simulation.}
  \label{fig:figure1}
  \vspace{-20pt}
\end{figure}

This semantic gap has direct consequences for downstream ITS tasks. Consider traffic simulation, which is widely used for safety testing and autonomous vehicle certification: existing language-driven simulators~\cite{zhong2023guided,zhong2023language,tan2023language,xia2024language} condition trajectory generation on text intent and raw HD map pixels, but they have no semantic understanding of the scene.  Generated scenarios therefore exhibit lane violations, ignored right-of-way rules, and unrealistic collisions, reducing their value for safety-critical validation. 

To address these limitations, We introduce \textbf{ChatBEV-QA}, a large-scale BEV VQA benchmark designed to cover the full range of traffic-relevant scene understanding tasks. An efficient automated construction pipeline generates 137K+ QA pairs from real-world driving datasets, spanning six question types across three dimensions: global understanding including area type and lane type, which determines the macro-level traffic context that governs vehicle behavior; vehicle-lane interactions such as lane centerline and navigation information, which underpin trajectory prediction and route planning in ITS; and vehicle-vehicle interactions (presence detection and relative orientation), which are essential for traffic conflict analysis and cooperative driving. All answers are generated by carefully designed rulebased functions to ensure label accuracy. Building on ChatBEV-QA, we fine-tune \textbf{ChatBEV}, a specialized VLM-based scene understanding model  that accurately interprets structured BEV map content directly from BEV images paired with natural language queries. ChatBEV-LLaVA-1.5 achieves the optimal balance across all six question types and generalizes well to unseen traffic datasets without additional fine-tuning. Finally, we design a novel \textbf{language-guided  traffic simulation framework} wherein ChatBEV serves as a specialized scene understanding module. By incorporating the structured scene understanding information extracted by ChatBEV including global scene context encoding road area types and navigation guidance providing lane centerlines, our framework synthesizes more realistic and map-compliant trajectories aligned with textual intent. Experiments on nuPlan demonstrate that leveraging this semantic knowledge consistently improves trajectory realism and substantially reduces scenario collision rates.

\section{Related Work}
\label{sec:related}
\subsection{Traffic Scene Understanding}
Traditional methods~\cite{li2020end, roddick2020predicting, can2022understanding} are generally task-specific, concentrating on discrete objectives like object detection, object recognition, or trajectory prediction. These approaches cannot answer free-form queries and lack the holistic reasoning needed for general applications. Recent research~\cite{deruyttere2019talk2car, vasudevan2018object, wu2023referring, wu2023language, choudhary2023talk2bev, cao2024maplm} has introduced vision-language models to traffic scenes to enable more flexible scene understanding. However, Talk2BEV~\cite{choudhary2023talk2bev} adapts VLMs to BEV understanding but is limited in scale and lacks vehicle-lane reasoning. MAPLM~\cite{cao2024maplm} focuses on road condition recognition using point clouds and language but omits vehicle interaction understanding. Our work fills this gap with a comprehensive BEV VQA benchmark and a specialized model covering all key traffic scene understanding dimensions.

\subsection{Language-guided Traffic Simulation}
Realistic traffic simulation is essential for ITS safety testing and autonomous vehicle validation. While rule-based simulators~\cite{erdmann2015sumo} offer controllability, they lack behavioral realism. Learning-based methods~\cite{tan2021scenegen, suo2021trafficsim} improve fidelity, and recent language-driven approaches~\cite{zhong2023language, xia2024language} use LLMs to translate natural language into generation objectives. However, existing models often ignore high-level map semantics, such as how lane structures and area contexts should constrain trajectories. We address this by integrating ChatBEV’s structured scene understanding into the simulation pipeline to ensure map-compliant and realistic vehicle behaviors.

\section{ChatBEV-QA: A BEV-based Traffic Scene Understanding Benchmark}
\label{sec:benchmark}

\subsection{Task Definition: Multi-Level Traffic Scene Understanding}
\label{sec:taskdef}

Effective traffic scene understanding from BEV maps requires reasoning at multiple levels that directly correspond to ITS operational needs. We define three complementary dimensions, each covered by two question types, so that the benchmark spans the full range of semantic queries.

\begin{itemize}[leftmargin=1em]
    \item \textbf{Global Understanding}: Traffic scene interpretation requires determining the environment type, as right-of-way norms, speed limits, and permitted maneuvers differ between intersections, highways, and parking areas. We use \textit{area type} to classify this environment and \textit{lane type} to identify lane function (e.g., straight, left-turn), providing the macro-level context that governs all subsequent behavioral judgments.
    \item \textbf{Vehicle-Lane Interaction}: It is critical to determine whether a vehicle occupies its designated lane and which lanes are consistent with its intended maneuver. We use \textit{location} to identify the lane currently occupied and \textit{navigation} to identify the lanes relevant to the next maneuver, enabling lane-level compliance reasoning that is central to ADAS monitoring functions.
    \item \textbf{Vehicle-Vehicle Interaction}: Assessing conflict risk and coordinating evasive responses depends on understanding the relative positions and headings of nearby vehicles. We use \textit{existence} to detect whether a vehicle is present in each surrounding direction and \textit{relative orientation} to describe its heading, capturing the spatial dynamics needed for conflict detection and cooperative driving analysis.
\end{itemize}

\subsection{Automated Data Construction Pipeline}

\begin{figure*}[t]
  \centering
  \subfloat[Two-stage automated ChatBEV-QA construction pipeline.\label{fig:pipeline}]{%
    \includegraphics[width=0.76\textwidth]{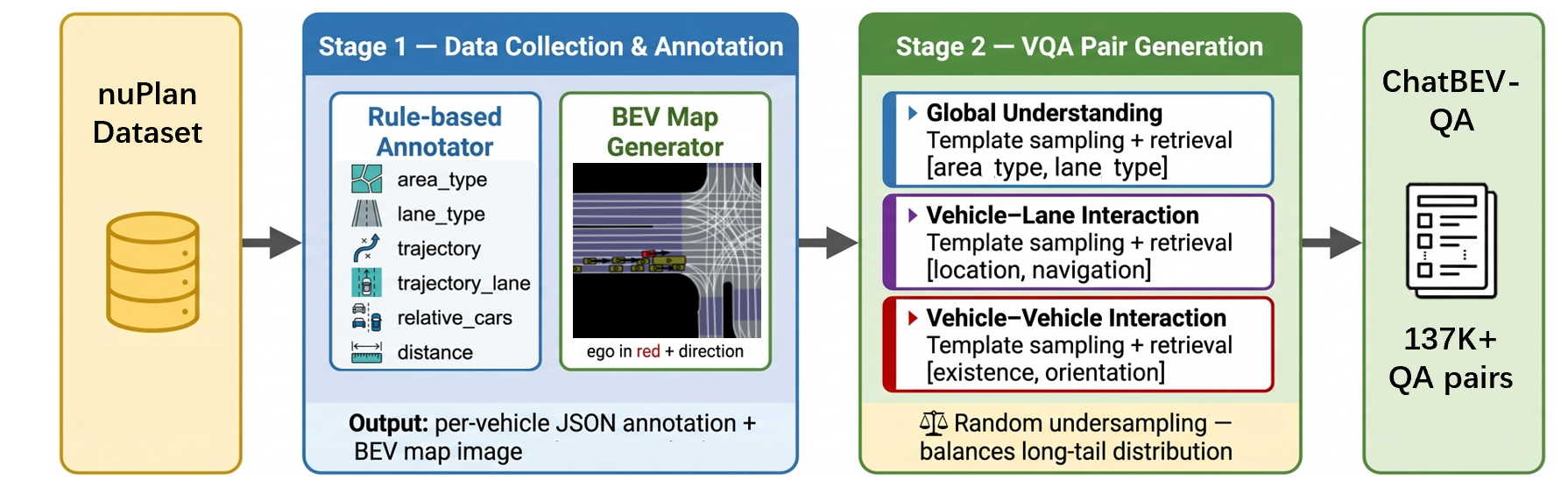}}%
  \hfill
  \subfloat[The VQA sample. \label{fig:sample}]{%
    \includegraphics[width=0.19\textwidth]{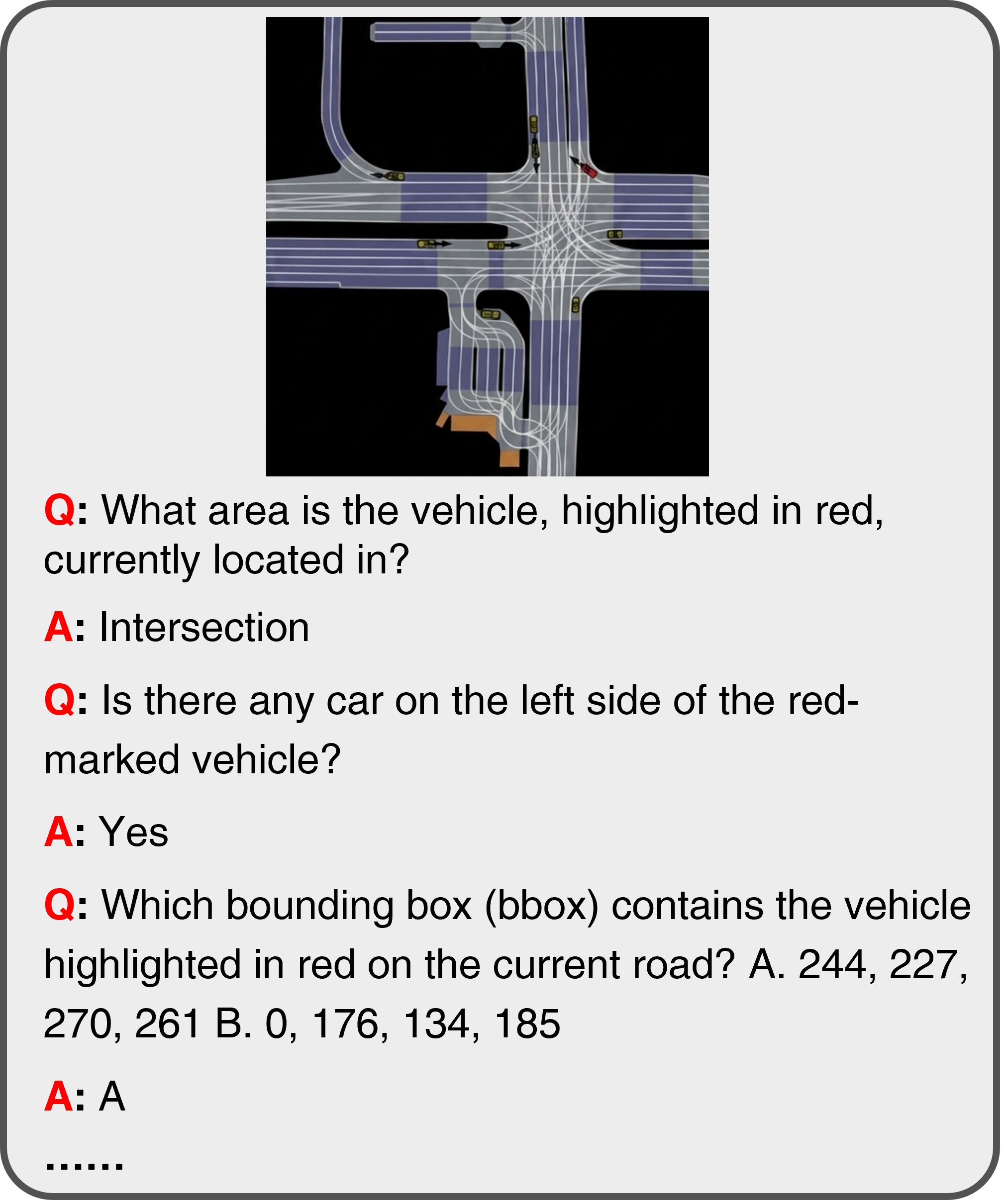}}
  \vspace{-4pt}
  \caption{Illustration of automated data construction pipeline and VQA sample.}
  \vspace{-10pt}
  \label{fig:stats}
\end{figure*}

Building on the question types defined above, we construct ChatBEV-QA from the nuPlan~\cite{caesar2021nuplan} dataset via a two-stage automated pipeline, as shown in Fig.~\ref{fig:pipeline}. This ensures that the dataset accurately reflects real-world traffic scenarios and constraints.


\subsubsection{Stage 1: Data Collection and Annotation}
Building on nuPlan's basic annotations of vehicle positions, speeds, and lane details, we develop a rule-based annotator validated via a human-in-the-loop workflow to extract high-level semantic fields per vehicle per timestep.  We then generate a JSON-formatted annotation using the designed functions for each vehicle in a scene at each timestep, containing the following fields: 1).~\texttt{area\_type}, indicating the current area type the vehicle is in; 2).~\texttt{lane\_type}, specifying the type of the lane the vehicle is currently in;  3).~\texttt{trajectory}, describing the behavioral categories corresponding to the vehicle's trajectory over the next 50 timesteps; 4).~\texttt{trajectory\_lane}, capturing the specific lane IDs traversed by the trajectory for the next 50 frames; 5).~\texttt{relative\_cars}, storing the IDs of other vehicles located in the four directions around the vehicle; 6).~\texttt{distance}, calculating the distances between the current vehicle and all other vehicles. 

For the BEV scene images, we develop a map generator and generate BEV maps for each vehicle in every scene, using the original map and trajectory annotations from nuPlan, with the coordinate origin placed at the ego vehicle's position. The vehicle of interest is highlighted in red to draw attention, with arrows indicating its direction if in motion. Our generated BEV map clearly marks lane boundaries and area divisions, providing the essential information for vehicle-lane interactions and global understanding.

\subsubsection{Stage 2: VQA Pair Generation}
The VQA generator randomly selects a question template and retrieves the answer from annotations. For \textit{area type} and \textit{lane type}, the annotation value is used directly. For \textit{location} and \textit{navigation}, two bounding-box choices are provided, one correct lane and one distractor from non-overlapping lanes. For \textit{existence} and \textit{relative orientation}, the generator selects a random direction and queries the spatial annotations. To mitigate the long-tail distribution inherent in the nuPlan dataset (e.g., predominant straight-lane scenario), we randomly undersample the majority class during construction to achieve a more balanced dataset. The example of the final VQA is shown in Fig.~\ref{fig:sample}.

\subsection{Dataset Statistics and Metrics}
\label{sec:stats}
\begin{wrapfigure}{r}{0.45\columnwidth} 
    \centering
    \vspace{-15pt} 
    \includegraphics[width=0.45\columnwidth]{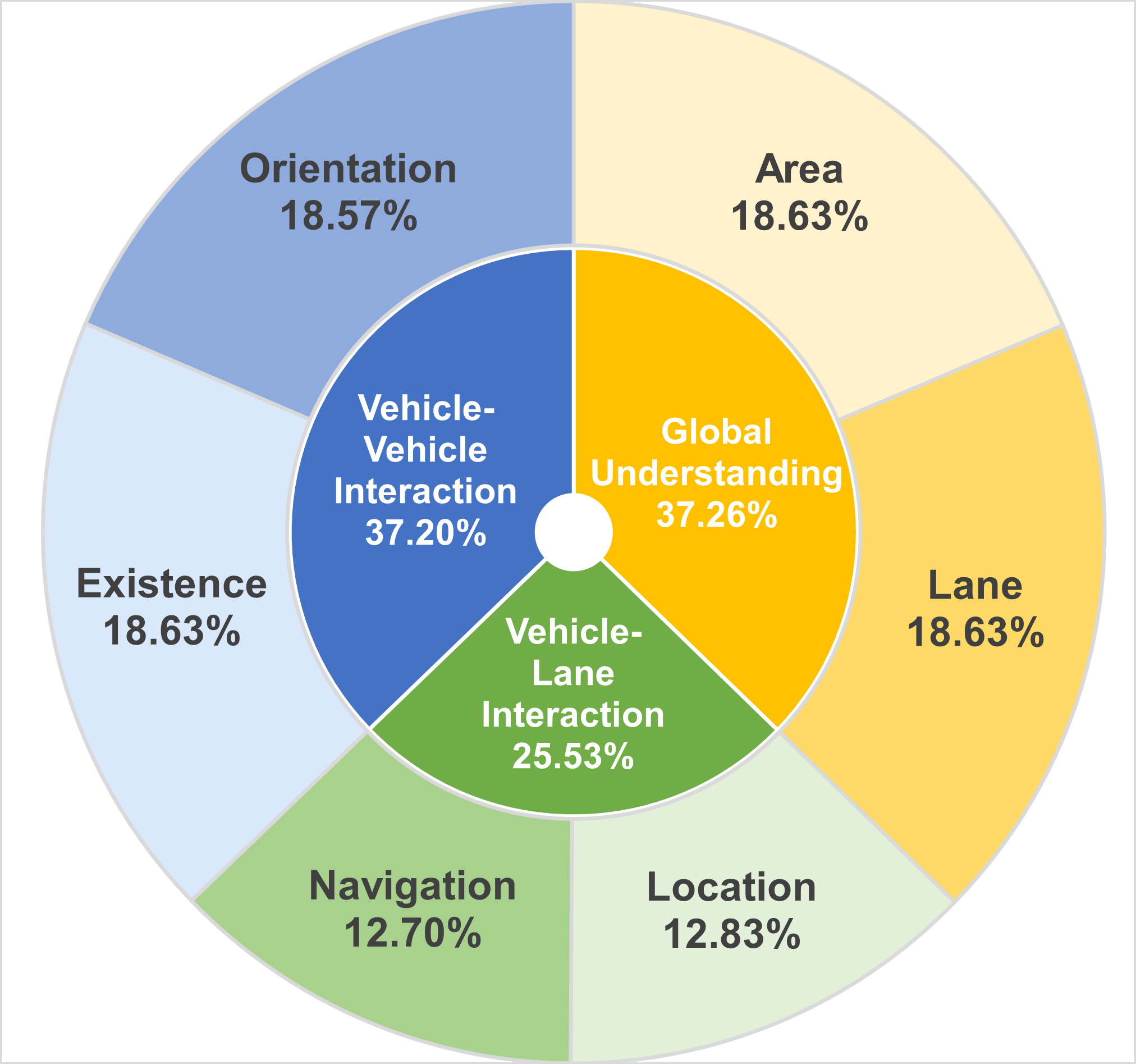}
    \caption{Question type dist.}
    \label{fig:data(b)}
    \vspace{-10pt} 
\end{wrapfigure}
\begin{figure}[htbp]
    \centering
    \includegraphics[width=0.9\columnwidth]{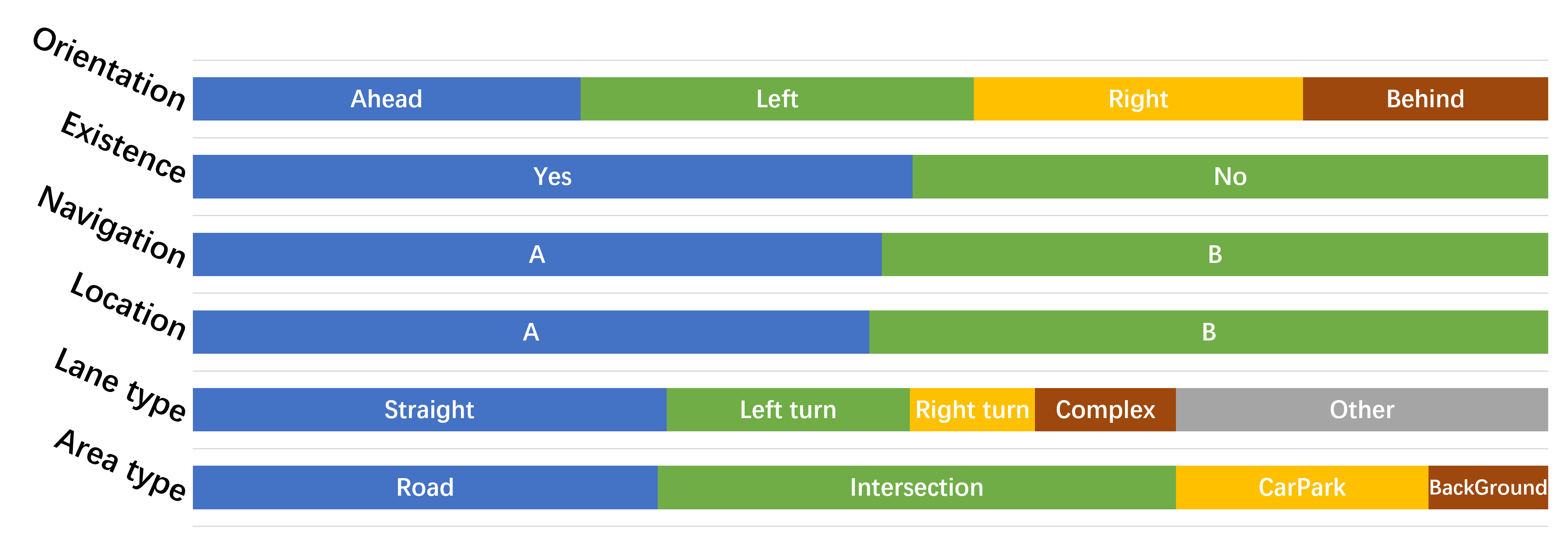}
    \caption{Answer category distribution in ChatBEV-QA.}
    \label{fig:data(c)}
    \vspace{-10pt}
\end{figure}

ChatBEV-QA comprises 137,818 QA pairs across 25,331 BEV images with 116,112 training samples and 21,706 test samples, averaging 5.44 questions per image. Fig.~\ref{fig:data(b)} and Fig.~\ref{fig:data(c)} shows balanced distributions across question types and answer categories. As summarized in Table~\ref{tab:dataset_comparison}, ChatBEV-QA significantly advances existing benchmarks in both scale and semantic depth. This high degree of automation ensures seamless scalability to full-scale repositories like nuPlan~\cite{caesar2021nuplan}, while our evaluation on nuScenes~\cite{nuscenes2019} and Lyft5~\cite{houston2020thousandhoursselfdrivingmotion} further demonstrates its robust generalization capabilities across diverse autonomous driving environments. Following VQA standards, we evaluate performance using top-1 accuracy, supplemented by a detailed breakdown by question type.

\begin{table}[ht]
\centering
\caption{Comparison of BEV VQA benchmarks. Under., Inter., V and L denote understanding, interaction, Vehicle and Lane, respectively.}
\vspace{-5pt}
\setlength{\tabcolsep}{3pt} 
\resizebox{\columnwidth}{!}{
\begin{tabular}{lccccc}
\hline
Dataset & Amount & Auto-pipeline & Global Under. & V-V Inter. & V-L Inter. \\ 
\hline
Talk2BEV~\cite{choudhary2023talk2bev} & 20K & $\times$ & \checkmark & \checkmark & $\times$ \\
MAPLM~\cite{cao2024maplm}    & 60K & $\times$ & \checkmark & $\times$ & $\times$ \\
\textbf{ChatBEV-QA} & \textbf{137K} & \checkmark & \checkmark & \checkmark & \checkmark \\
\hline
\end{tabular}
}
\vspace{-10pt}
\label{tab:dataset_comparison}
\end{table}

\subsection{ChatBEV: VLM-based Traffic Scene Understanding Module}
\label{sec:chatbev}

ChatBEV is a VLM fine-tuned on ChatBEV-QA via LoRA visual instruction tuning. Given a BEV image and a natural language question from one of the six question types defined above, it produces a text answer describing the corresponding semantic attribute of the scene. Each fine-tuned variant is denoted ChatBEV-[BaseModel]. The model is trained end-to-end on paired (BEV image, question, answer) examples from ChatBEV-QA, learning to interpret BEV-specific visual patterns such as lane boundaries, road area divisions, and the spatial arrangement of vehicles. As shown in Table~\ref{tab:base_VLM}, \textbf{ChatBEV-LLaVA-1.5} achieves the best overall balance across all six question types and is used as the primary model in downstream applications.

ChatBEV provides robust scene understanding capabilities that directly benefit downstream applications. An intersection controller can query ``Is there a vehicle approaching from the left?'' and receive a reliable answer for conflict assessment. A traffic analyst can interrogate a stored BEV feed in natural language to audit right-of-way behavior across thousands of scenarios. In traffic simulation, ChatBEV can translate BEV images into structured semantic constraints, enabling realistic vehicle behaviors that comply with traffic rules.

\begin{figure*}[t]
  \centering
  \includegraphics[width=0.75\linewidth]{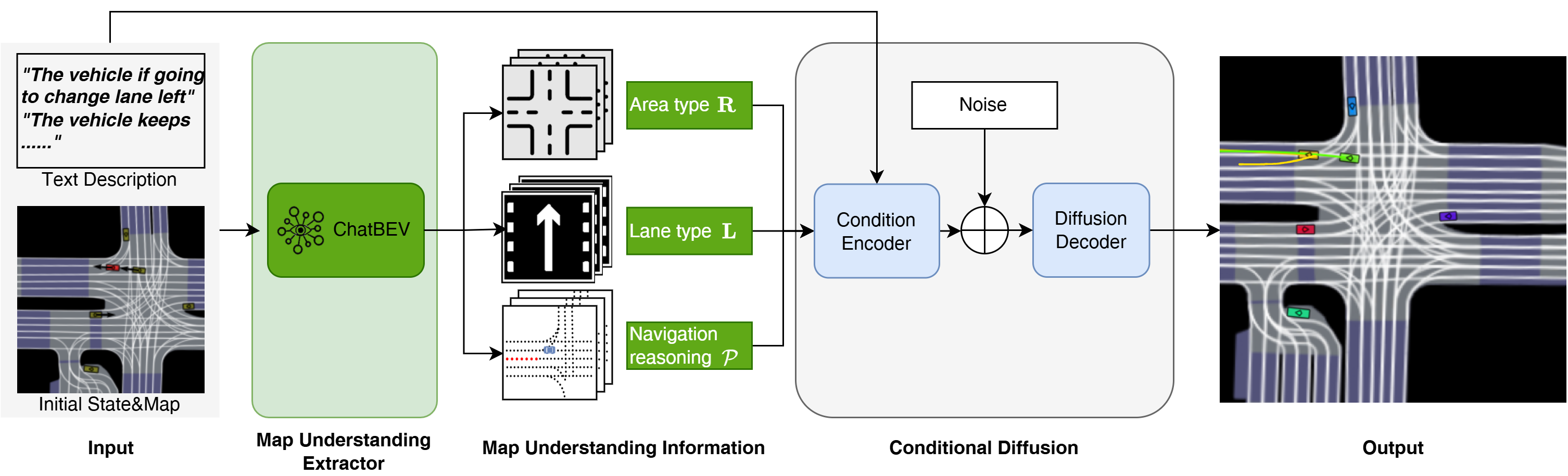}
  \vspace{-5pt}
  \caption{Inference pipeline of the language-driven traffic simulation framework. ChatBEV extracts structured semantic map understanding that conditions the diffusion decoder for text-consistent, map-compliant trajectory synthesis.}
  \label{fig:scen_gen}
  \vspace{-12pt}
\end{figure*}

\section{Application on Language-guided Traffic Simulation}
\label{sec:generation}

Language-guided traffic simulation is an important downstream application where ChatBEV's semantic scene understanding directly benefits system performance. Existing language-driven simulators~\cite{zhong2023language, xia2024language} rely on raw maps without semantic interpretation, failing to enforce lane-following or traffic rule compliance. We address this by integrating ChatBEV as a scene understanding module (Fig.~\ref{fig:scen_gen}). For each vehicle, ChatBEV translates BEV images into structured semantic representations, such as area types and navigation paths, which serve as high-level constraints for the trajectory generator alongside textual descriptions.

\subsection{Problem Formulation}

We formulate the scene generation task as an imitation learning problem following ~\cite{zhong2023guided, zhong2023language}. Given a scene with $N$ vehicles, we define the states of all vehicle over $T$ timesteps as $\mathbb{S}=[\mathbf{S}_1, \mathbf{S}_2, \cdots, \mathbf{S}_N]\in \mathbb{R}^{N\times T\times 4}$, where $\mathbf{S}_i=[s_i^1, s_i^1, \cdots, s_i^T]$ and $s_i^t=(x_i^t, y_i^t, v_i^t, \theta_i^t)$ denotes the current state~(2D location, speed, and yaw) of vehicle $i$ at the $t$-th timestep. Similarly, we can define the corresponding actions $\mathbb{A}\in \mathbb{R}^{N\times T\times 2}$, here $\mathbf{A}_i=[a_i^0, a_i^1, \cdots, a_i^{T-1}]$ and each $a_i^t=(\dot{v}_i^t, \dot{\theta}_i^t)$ is the action~(acceleration and yaw rate) of vehicle $i$ at the $t$-th timestep. Future states are determined by unicycle dynamic model $f$ with $s_i^{t+1}=f(s_i^t, a_i^t)$. The overall trajectory is $\boldsymbol{\tau}=[\mathbb{A}, \mathbb{S}]$. 
To generate future trajectories, the model conditions on  $C=(I, D, \mathbb{S}^h, M)$ to be decision-relevant context encompassing raw map data $I$, text descriptions $D$, history $\mathbb{S}^h=[\mathbf{S}^{-H}, \cdots, \mathbf{S}^0]$ , and ChatBEV-inferred structured semantic context $M$.

\subsection{Scene Understanding Integration}
\label{sec:sim_integration}

Given the initial state of all vehicles and the corresponding textual description, we first extract semantic map information $M=(\mathbf{V},\mathcal{P})$  and integrate it with the original input to guide the scene generation module. At inference time, ChatBEV is queried with the BEV image and question prompts for each vehicle. Its answers are parsed to derive the semantic map information. During training, these information is derived directly from ground-truth annotations to eliminate noise.

The \textbf{global scene context} $\mathbf{V}=[\mathbf{R},\mathbf{L}]\in\mathbb{R}^{N\times9}$ is derived from ChatBEV's answers to the area type and lane type questions. The predicted class labels are encoded as one-hot vectors, with area type $\mathbf{R}\in\mathbb{R}^{N\times4}$ and lane type $\mathbf{L}\in\mathbb{R}^{N\times5}$, encoding the macro-level behavioral rules of the scene (e.g., yielding norms at intersections).

The \textbf{navigation guidance} $\mathcal{P}\in\mathbb{R}^{N\times N_s\times N_p\times d}$ is derived from ChatBEV's answer to the navigation question, which predicts a bounding box over the lanes most consistent with the text description. All lane centerlines within this bounding box are collected as $\mathcal{P}$, where $N_s$, $N_p$, and $d$ are the number of lanes, points per centerline, and point dimension. This constrains the generated trajectory to lanes that are functionally aligned with the intended maneuver.

\begin{table*}[!t]
\small
\caption{Experiment results on ChatBEV-QA. Best results in \textbf{bold}. ChatBEV achieves significant improvements across all aspects.}
\begin{center}
\resizebox{0.85\textwidth}{!}{
\begin{tabular}{l|cc|cc|cc|c}
\toprule
 \multirow{2}{*}{Model} & \multicolumn{2}{c}{Global Understanding} & \multicolumn{2}{|c}{Vehicle-Lane Interaction} & \multicolumn{2}{|c|}{Vehicle-Vehicle Interaction} & \multirow{2}{*}{Overall} \\
\cmidrule{2-7}
 & Area & Lane & Location & Navigation & Existence & Orientation & \\
\midrule
MCAN~\cite{yu2019deepmodularcoattentionnetworks} & 0.359 & 0.316 & 0.505 & 0.492 & 0.568 & 0.267 & 0.410 \\ MMNasnet~\cite{yu2020deep} & 0.358 & 0.311 & 0.501 & 0.522 & 0.615 & 0.252 & 0.418 \\
BUTD~\cite{anderson2018bottomuptopdownattentionimage} & 0.360 & 0.310 & 0.511 & 0.725 & 0.602 & 0.269 & 0.448 \\
ConvNeXt~\cite{liu2022convnet2020s} & 0.896 & \textbf{0.708} & - & - & 0.644 & 0.304 & - \\
ViT~\cite{dosovitskiy2021imageworth16x16words} & 0.878 & 0.707 & - & - & 0.556 & 0.321 & - \\
 LXMERT~\cite{tan2019lxmert} & 0.793 & 0.703 & 0.499 & 0.478 & 0.728 & 0.335 & 0.599 \\
UNITER~\cite{chen2020uniter} & 0.748 & 0.672 & 0.499 & 0.478 & 0.790 & 0.373 & 0.604 \\
 GPT-4~\cite{achiam2023gpt} & 0.441 & 0.407 & 0.372 & 0.438 & 0.476 & 0.239 & 0.395 \\
 LLaVA-1.5~\cite{liu2024visual} & 0.393 & 0.406 & 0.543 & 0.549 & 0.514 & 0.240 & 0.431 \\
 Gemini~\cite{team2023gemini} & 0.465 & 0.452 & 0.572 & 0.686 & 0.532 & 0.234 & 0.477 \\
ChatBEV & \textbf{0.908} & 0.706 & \textbf{0.800} & \textbf{0.805} & \textbf{0.866} & \textbf{0.469} & \textbf{0.754} \\
\bottomrule
\end{tabular}}
\end{center}
\vspace{-10pt}
\label{tab:main_table}
\end{table*}

\begin{table*}[!t]
\small
\begin{center}

\caption{Impact of different base VLMs. Best in \textbf{bold}, second best \underline{underlined}. ChatBEV-LLaVA-1.5 yields the best overall performance.}

\label{tab:base_VLM}
\vspace{-5pt}
\resizebox{0.95\textwidth}{!}{
\begin{tabular}{l|cc|cc|cc|c}
\toprule
\multirow{2}{*}{Model} & \multicolumn{2}{c}{Global Understanding} & \multicolumn{2}{|c}{Vehicle-Lane Interaction} & \multicolumn{2}{|c|}{Vehicle-Vehicle Interaction} & \multirow{2}{*}{Overall} \\
\cmidrule{2-7}
& Area & Lane & Location & Navigation & Existence & Orientation & \\
\midrule
ChatBEV-Qwen2.5-VL & 0.859 & 0.421 & 0.783 & \underline{0.797} & 0.617 & 0.329 & 0.620 \\
ChatBEV-InternLM-XComposer2 & 0.903 & 0.300 & \textbf{0.920} & 0.795 & 0.710 & 0.261 & 0.629 \\
ChatBEV-InternVL2-5 & 0.787 & 0.533 & \textbf{0.920} & 0.509 & 0.629 & 0.442 & 0.630 \\
ChatBEV-BLIP & 0.880 & \textbf{0.754} & 0.501 & 0.522 & 0.789 & 0.425 & 0.658 \\
ChatBEV-LLaVA-1.5 & \textbf{0.908} & \underline{0.706} & \underline{0.800} & \textbf{0.805} & \textbf{0.866} & \textbf{0.469} & \textbf{0.754} \\
\bottomrule
\end{tabular}
}
\end{center}
\vspace{-10pt}
\end{table*}

\begin{table*}[!t]
\small
\begin{center}
\caption{Generalization on nuScenes and Lyft5. Best results in \textbf{bold}. ChatBEV shows strong zero-shot generalization across datasets.}
\label{tab:generalization}
\vspace{-5pt}
\resizebox{0.95\textwidth}{!}{
\begin{tabular}{l|cccc|c||cccc|c}
\toprule
\multirow{2}{*}{Model}
& \multicolumn{5}{c||}{nuScenes}
& \multicolumn{5}{c}{Lyft5} \\
\cmidrule(lr){2-6} \cmidrule(lr){7-11}
& Area & Lane & Existence & Orientation & Overall
& Area & Lane & Existence & Orientation & Overall \\
\midrule
MCAN~\cite{yu2019deepmodularcoattentionnetworks} & 0.279 & - & 0.347 & 0.268 & 0.316 & 0.494 & 0.611 & 0.609 & 0.312 & 0.507 \\
MMNasnet~\cite{yu2020deep} & 0.259 & - & 0.352 & 0.222 & 0.307 & 0.458 & 0.573 & 0.649 & 0.289 & 0.493 \\
BUTD~\cite{anderson2018bottomuptopdownattentionimage} & 0.244 & - & 0.362 & 0.206 & 0.291 & 0.494 & 0.590 & 0.617 & 0.300 & 0.501 \\
LXMERT~\cite{tan2019lxmert} & 0.639 & - & 0.639 & 0.270 & 0.516 & 0.724 & 0.736 & 0.743 & 0.359 & 0.641 \\
UNITER~\cite{chen2020uniter} & 0.622 & - & 0.598 & 0.245 & 0.488 & 0.667 & 0.689 & 0.628 & 0.390 & 0.594 \\
GPT-4~\cite{achiam2023gpt} & 0.234 & - & 0.532 & 0.278 & 0.348 & 0.524 & 0.608 & 0.523 & 0.346 & 0.501 \\
LLaVA-1.5~\cite{liu2024visual} & 0.061 & - & 0.540 & 0.253 & 0.285 & 0.448 & 0.655 & 0.498 & 0.283 & 0.471 \\
Gemini~\cite{team2023gemini} & 0.098 & - & 0.540 & 0.245 & 0.294 & 0.588 & 0.667 & 0.545 & 0.341 & 0.536 \\
\textbf{ChatBEV} & \textbf{0.791} & - & \textbf{0.702} & \textbf{0.347} & \textbf{0.613}
& \textbf{0.782} & \textbf{0.757} & \textbf{0.889} & \textbf{0.574} & \textbf{0.751} \\
\bottomrule
\end{tabular}
}
\end{center}
\vspace{-10pt}
\end{table*}

\begin{table}[h]

\caption{Impact of the noisy BEV Map.}
\vspace{-10pt}

\label{tab:noise_level}
\begin{center}
\begin{tabular}{l|cccc}
\toprule
Noise                    & 0\% & 10\% & 20\% & 30\%  \\
\midrule
Lane Disruption Noise & 0.754 & 0.742 & 0.735 & 0.727 \\
Vehicle Disruption Noise & 0.754 & 0.741 & 0.736 & 0.727 \\
Combined Noise & 0.754 & 0.740 & 0.734 & 0.723 \\

\bottomrule
\end{tabular}

\end{center}
\vspace{-20pt}

\end{table}
\subsection{Conditional Diffusion Architecture}
The conditional diffusion architecture consists of a condition encoder $f_{\mathrm{enc}}$ that maps heterogeneous inputs, including historical states $\mathbb{S}^h$, raw map data $I$, text descriptions $D$, and structured map understanding information $M=(\mathbf{V}, \mathcal{P})$, into a unified embedding $\mathbb{E} \in \mathbb{R}^{N\times H\times d_h}$ via dedicated feed-forward modules. Subsequently, the diffusion decoder $f_{\theta}$ iteratively denoises a random trajectory $\mathbb{A}_K$ conditioned on $\mathbb{E}$ to produce the final action sequence $\mathbb{A}$, which is then converted into a physically feasible state trajectory $\mathbb{S}$ through unicycle dynamics rollout, yielding the complete pipeline: $\boldsymbol{\tau} \;=\; f_{\theta}\!\left(\mathbb{A}_K,\; f_{\mathrm{enc}}(\mathbb{S}^h,I, D, M)\right)$

We validate this pipeline on two architecturally distinct decoders. CTG~\cite{zhong2023guided} is a per-vehicle model using a U-Net with temporal 1D convolutions. CTG++~\cite{zhong2023language} models all $N$ vehicles jointly via a spatio-temporal transformer and map attention at each denoising step. Consistent gains across both decoders confirm that ChatBEV's benefit is backbone-agnostic.

\section{Experiments}
\label{sec:experiments}

\subsection{Evaluation on ChatBEV-QA}

\subsubsection{Baselines}

We compare ChatBEV against several baselines categorized by their approach to traffic scenario understanding: 1).~\textbf{Traditional Scene Interpreters}: including traditional deep learning-based VQA models~\cite{yu2019deepmodularcoattentionnetworks,yu2020deep,anderson2018bottomuptopdownattentionimage}, pre-trained VQA models\cite{chen2020uniter,tan2019lxmert} and vision models\cite{dosovitskiy2021imageworth16x16words,liu2022convnet2020s}. To ensure a fair comparison with methods~\cite{yu2019deepmodularcoattentionnetworks,yu2020deep,anderson2018bottomuptopdownattentionimage,chen2020uniter,tan2019lxmert} relying on extracted features rather than raw images, we adopt a consistent pipeline to process our generated BEV images. For vision models, tasks are reformulated as classification problems, though vehicle–lane interaction tasks are omitted due to incompatibility with this format. 2).~\textbf{VLM-based Interpreters}: While the most relevant work, Talk2BEV~\cite{choudhary2023talk2bev} and MAPLM~\cite{cao2024maplm}are not publicly available for direct benchmarking, we establish ChatBEV as an open-source baseline for the community. To ensure a rigorous evaluation, we instead compare our framework against several state-of-the-art general-purpose VLMs such as \cite{achiam2023gpt,liu2024visual,team2023gemini}.

\subsubsection{Quantitative Results}
Table~\ref{tab:main_table} shows that traditional VQA models struggle to interpret BEV maps accurately. Pretrained vision models such as ConvNeXt and ViT handle simple perception but fail at relational reasoning requiring spatial understanding. VQA-based models such as LXMERT and UNITER better capture vehicle interactions but remain suboptimal on lane-interaction tasks. Zero-shot VLMs perform comparably to traditional trained models, indicating a substantial domain gap. \textbf{ChatBEV significantly outperforms all baselines} after fine-tuning, confirming the value of both the benchmark and the specialized training approach.

\subsubsection{Impact of Base VLM}
As shown in Table~\ref{tab:base_VLM}, while Qwen2.5-VL, InternLM-XComposer2, InternVL2-5, and BLIP excel on specific subtasks, ChatBEV-LLaVA-1.5 delivers the most balanced and consistent performance across all six question types and is adopted as our primary model.

\begin{table*}[!t]
\small
\caption{Generation results with different inputs. ChatBEV-inferred map understanding greatly enhances trajectory realism.}
\begin{center}
\resizebox{0.88\textwidth}{!}{
\begin{tabular}{l|l|ccccc}
\toprule
Decoder & Input & mADE~$\downarrow$ & minADE~$\downarrow$ & mFDE~$\downarrow$ & minFDE~$\downarrow$ & SCR~$\downarrow$ \\
\midrule
\multirow{2}{*}{CTG++} & w/ $D$         & 1.582 & 0.867 & 3.689 & 1.922 & 0.029 \\
                        & w/ $D\,{\&}\,M$  & 1.293~$(\downarrow18.3\%)$ & 0.810~$(\downarrow6.6\%)$ & 2.943~$(\downarrow20.2\%)$ & 1.705~$(\downarrow11.3\%)$ & 0.018~$(\downarrow37.9\%)$ \\
\midrule
\multirow{2}{*}{CTG}    & w/ $D$         & 1.246 & 0.752 & 3.247 & 1.898 & 0.022 \\
                        & w/ $D\,{\&}\,M$  & 0.985~$(\downarrow20.9\%)$ & 0.747~$(\downarrow0.7\%)$ & 2.456~$(\downarrow24.4\%)$ & 1.810~$(\downarrow4.6\%)$ & 0.020~$(\downarrow9.1\%)$ \\
\bottomrule
\end{tabular}
}
\end{center}
\vspace{-10pt}
\label{tab:CTGplus}
\end{table*}

\begin{figure*}[t]
  \centering
  \includegraphics[width=0.75\linewidth]{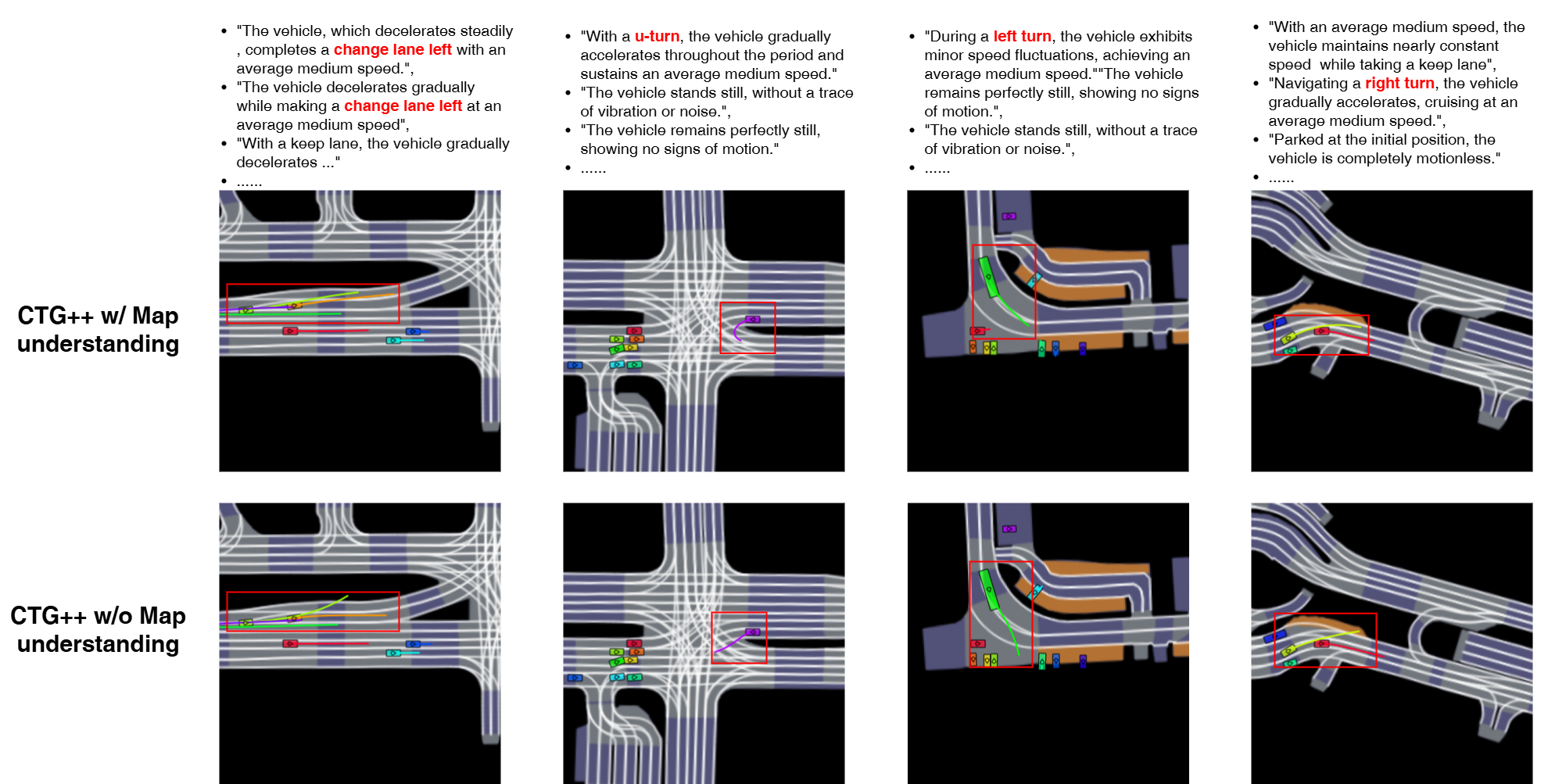}
  \vspace{-5pt}
  \caption{Qualitative generation comparisons. ChatBEV map understanding reduces off-lane driving and enables correct handling of rare maneuvers such as the U-turn shown in column 2.}
  \label{fig:vis}
  \vspace{-15pt}
\end{figure*}

\begin{table}[!t]
\small

\caption{Impact of map information. Best results are in \textbf{bold}. Global understanding and navigation reasoning both improve performance, with their combination performing best.}
\label{tab:ablation_mapinfo}
\vspace{-10pt}
\begin{center}
\setlength{\tabcolsep}{4pt}
\begin{tabular}{l|cccc}
\toprule
Map Information                     & mADE & minADE & mFDE & minFDE \\
\midrule
- & 1.582    & 0.867    & 3.689    & 1.922\\
only $\mathcal{P}$ & 1.241 & 0.695 & 2.643 & 1.357 \\
only $\mathbf{V}$   & 1.569 & 0.829 & 3.596 & 1.741 \\
$\mathcal{P}+\mathbf{V}$                     & \textbf{1.143} & \textbf{0.679} & \textbf{2.505} & \textbf{1.315} \\
\bottomrule
\end{tabular}

\end{center}
\vspace{-20pt}
\end{table}

\subsubsection{Generalization to Unseen Scenes}
A key requirement for ITS deployment is robust operation across diverse road networks and sensor configurations without per-city retraining. To assess this, we apply our automated pipeline to the nuScenes and Lyft5 datasets, generating approximately $1,000$ test samples for each. Table~\ref{tab:generalization} demonstrates that ChatBEV consistently outperforms all baselines in a zero-shot setting, showcasing strong transferability and eliminating the need for costly per-site retraining in new environments.

\subsubsection{Robustness to Perception Noise}
Real-world perception is often imperfect due to occlusions, adverse weather, and localization drift. To simulate these conditions, we introduce three noise types during BEV generation: Vehicle Disruption (positional perturbation or omission), Lane Disruption (degraded markings), and Combined Noise. Table~\ref{tab:noise_level} shows that while performance declines with noise, ChatBEV sustains satisfactory accuracy even at a 30\% noise level, confirming its robustness under the imperfect perception conditions characteristic of real traffic infrastructure.

\subsection{Language-guided Traffic Simulation}

\subsubsection{Setup}
We train and evaluate on the nuPlan mini dataset, utilizing rule-based functions to generate text instructions (e.g., ``aggressive lane change''). Scenario realism is assessed via Mean/Minimum Average Displacement Error (mADE/minADE), Final Displacement Error (mFDE/minFDE), and the Scenario Collision Rate (SCR), which directly measures the safety compliance of the generated scenarios.

\subsubsection{Impact of ChatBEV Scene Understanding}
Table~\ref{tab:CTGplus} highlights the improvement in simulation quality when ChatBEV provides semantic scene context. Incorporating ChatBEV's structured understanding yields consistent gains across both decoders: up to 20.9\% reduction in mADE with CTG and a substantial 37.9\% reduction in SCR with CTG++. For simulation engineers, this means generated test scenarios now respect real road semantics, including area-type constraints and lane-level compliance, producing higher-quality safety-critical test cases for autonomous vehicle certification and ITS validation pipelines.

\subsubsection{Ablation Study on Map Information}
Table~\ref{tab:ablation_mapinfo} isolates the effects of global scene context and lane-level navigation guidance with CTG decoder. To ensure reliable evaluation and eliminate perception noise, ground-truth annotations are utilized to derive $\mathbf{V},\mathcal{P}$ during training. Results indicate that navigation guidance offers the primary contribution by constraining the feasible trajectory space, while global context provides necessary complementary behavioral norms. Their combination yields the best results, proving that both scene-level and lane-level understanding are necessary for simulating realistic traffic scene.

\subsubsection{Qualitative Analysis of Interactive Scenarios}
Fig.~\ref{fig:vis} presents qualitative comparisons. Incorporating ChatBEV's interpretation yields trajectories that better respect lane boundaries and social conventions (e.g., maintaining safe gaps). Notably, in the U-turn case (column 2), a rare corner case is correctly executed only when high-level semantic reasoning is provided, demonstrating that ChatBEV supplies the structural grounding needed for challenging maneuvers.

\section{Conclusion}
\label{sec:conclusion}
In this paper, We introduce ChatBEV-QA, a scalable BEV VQA benchmark that covers a wide range of scene understanding tasks. Building on this, our ChatBEV model excels in comprehensive scene understanding and can provides advanced guidance for more controllable multi-agent traffic scene generation.
\textbf{Limitations}: The current scope focuses on vehicle dynamics, excluding pedestrians and cyclists, which remains a subject for future research.

\bibliographystyle{IEEEtran}
	\bibliography{ref} 

\end{document}